\pdfoutput=1 
\documentclass[11pt,a4paper]{article}
\usepackage[hyperref]{acl2020}
\usepackage{times}
\usepackage{latexsym}
\usepackage{booktabs}

\usepackage{microtype}

\aclfinalcopy 
\def\aclpaperid{1815} 

\usepackage{url}
\usepackage{graphicx}
\usepackage{adjustbox}
\usepackage{subcaption}
\usepackage{xspace}
\usepackage{comment}
\usepackage{amsmath}
\usepackage{xcolor}
\usepackage{multirow}

\graphicspath{../emojis/} 

\title{Simple, Interpretable and Stable Method \\ for Detecting Words with Usage Change across Corpora}

\author{Hila Gonen\thanks{~~Equal contribution.} $^1$ \space \space
  Ganesh Jawahar$^{*2}$ \space \space  Djam\'e Seddah$^{2}$ \space
  \space Yoav Goldberg$^{1,3}$ \\
	[0.5em]$^1$Department of Computer Science,  Bar-Ilan University \\
	$^2$Inria Paris\\
	$^3$Allen Institute for Artificial Intelligence \\
        {\tt\small hilagnn@gmail.com \quad ganeshjwhr@gmail.com}\\
         {\tt\small yoav.goldberg@gmail.com \quad djame.seddah@inria.fr}}

\date{}

\usepackage{fancyhdr}
\fancypagestyle{firststyle}
{
    \fancyhf{}
    \fancyhead{}
    \lhead{Published in ACL 2020}
}

\begin{document}
\maketitle
\thispagestyle{firststyle}
\begin{abstract}
    
    The problem of comparing two bodies of text and searching for words that differ in their usage between them arises often in digital humanities and computational social science. This is commonly approached by training word embeddings on each corpus, aligning the vector spaces, and looking for words whose cosine distance in the aligned space is large. However, these methods often require extensive filtering of the vocabulary to perform well, and---as we show in this work---result in unstable, and hence less reliable, results. We propose an alternative approach that does not use vector space alignment, and instead considers the neighbors of each word. The method is simple, interpretable and stable. We demonstrate its effectiveness in 9 different setups, considering different corpus splitting criteria (age, gender and profession of tweet authors, time of tweet) and different languages (English, French and Hebrew).

\end{abstract}

\section{Introduction}

Analyzing differences in corpora from different sources (different time periods, populations, geographic regions, news outlets, etc) is a central use case in digital humanities and computational social science. A particular methodology is to identify individual words that are used differently in the different corpora. This includes words that have their meaning changed over time periods \cite{KCHHP14,KAP15,HLJ16,KOSV18,NLA18}, and words that are used differently by different populations \cite{ADB17,RRAB17}.  It is thus desired to have an \emph{automatic}, \emph{robust} and \emph{simple} method for detecting such potential changes in word usage and surfacing them for human analysis. In this work we present such a method.

A popular method for performing the task (\S\ref{sec:align}) is to train word embeddings on each corpus and then to project one space to the other using a \emph{vector-space alignment} algorithm. Then, distances between a word-form to itself in the aligned space are used as an estimation of word usage change \cite{HLJ16}. We show that the common alignment-based approach is unstable, and hence less reliable for the usage change detection task (\S\ref{sec:stability},\S\ref{sec:experiments}). In addition, it is also sensitive to proper nouns and requires filtering them.

We propose a new and simple method for detecting usage change, that does not involve vector space alignment (\S\ref{sec:proposal}). Instead of trying to align two \emph{different} vector spaces, we propose to work directly in the \emph{shared} \emph{vocabulary space}: we take the neighbors of a word in a vector space to reflect its usage, and consider words that have drastically different neighbours in the spaces induced by the different corpora to be words subjected to usage change. 
The intuition behind this approach is that words that are used significantly differently across corpora are expected to have different contexts and thus to have only few neighboring words in common. In order to determine the extent of the usage change of a word, we simply consider its top-k neighbors in each of the two corpora, and compute the size of the intersection of the two lists. The smaller the intersection is, the bigger we expect the change to be. The words are ranked accordingly.

The advantages of our method are the following:

\begin{enumerate}
    \item \textbf{Simplicity}: the method is extremely simple to implement and apply, with no need for space alignment, hyperparameter tuning, and vocabulary filtering, except for simple frequency cutoffs. 
    \item \textbf{Stability}: Our method is stable, outputting similar results across different word embeddings trained on the same corpora, in contrast to the alignment-based approach.
    \item \textbf{Interpretability}: The ranking produced by our method is very intuitive to analyze. Looking at the neighborhood of a word in the two corpora reveals both the meaning of the word in each, and the extent to which the word has changed.
    \item \textbf{Locality}: The interpretability aspect is closely linked to the \emph{locality} of the decision. In our approach, the score of each word is determined only by its own neighbours in each of the spaces. In contrast, in the projection based method the similarity of a pair of words after the projection depends on the projection process, which implicitly takes into account \emph{all the other words in both spaces and their relations to each other}, as well as the projection lexicon itself, and the projection algorithm. This makes the algorithmic predictions of the projection-based methods opaque and practically impossible to reason about.
\end{enumerate}

We demonstrate the applicability and robustness of the proposed method (\S\ref{sec:experiments}) by performing a series of experiments in which we use it to identify word usage changes in a variety of corpus pairs, reflecting different data division criteria.  We also demonstrate the cross-linguistic applicability of the method by successfully applying it to two additional languages beyond English: French (a Romance language) and Hebrew (a Semitic language).

We argue that future work on detecting word change should use our method as an alternative to the now dominant projection-based method. To this end, we provide a toolkit for detecting and visualizing word usage change across corpora.\footnote{\url{https://github.com/gonenhila/usage_change}}

\section{Task Definition}

Our aim is to analyze differences between corpora by detecting words that are used differently across them. This task is often referred to as ``detecting meaning change" \cite{ADB17,DFB19}.

However, we find the name ``meaning change" to be misleading. Words may have several meanings in the different corpora, but different dominant sense in each corpus, indicating different use of the word. For this reason, we refer to this task as ``detecting usage change".

We define our task as follows: given two corpora with substantial overlapping vocabularies, identify words that their predominant use is different in the two corpora. The algorithm should return a ranked list of words, from the candidate that is most likely to have undergone usage-change, to the least likely.

Since the primary use of such algorithm is corpus-based research, we expect a human to manually verify the results. To this end, while the method does not need to be completely accurate, it is desirable that most of the top returned words are indeed those that underwent change, and it is also desirable to provide explanations or interpretations as to the usage of the word in each corpus. Lastly, as humans are susceptible to be convinced by algorithms, we prefer algorithms that reflect real trends in the data and not accidental changes in environmental conditions.

\section{Stability}
\label{sec:stability}
A desired property of an analysis method is \emph{stability}: when applied several times with slightly different conditions, we expect the method to return the same, or very similar, results. Insignificant changes in the initial conditions should result in insignificant changes in the output.  This increases the likelihood that the uncovered effects are real and not just artifacts of the initial conditions.

Recent works question the stability of word embedding algorithms, demonstrating that different training runs produce different results, especially with small underlying datasets. \newcite{AM18} focuses on the cosine-similarity between words in the learned embedding space, showing large variability upon minor manipulations on the corpus. \newcite{WKM18} make a similar argument, showing that word embeddings are unstable by looking at the 10-nearest neighbors (NN) of a word across the different embeddings, and showing that larger lists of nearest neighbors are generally more stable.

In this work, we are concerned with the stability of usage-change detection algorithms, and present a metric for measuring this stability. A usage-change detection algorithm takes as input two corpora, and returns a ranked list $r$ of candidate words, sorted from the most likely to have changed to the least likely. For a stable algorithm, we expect different runs to return similar lists. While we do not care about the exact position of a word within a list, we do care about the composition of words at the top of the list.
We thus propose a measure we call \textbf{intersection@$k$}, measuring the percentage of shared words in the the top-k predictions of both outputs:
\begin{equation}
\mathrm{intersection}@k(r_1, r_2) = \frac{| r^k_1 \cap r^k_2 |}{k}
\end{equation}
where $r_1$ and $r_2$ are the two ranked lists, and $r^k_i$ is the set of top $k$ ranked words in ranking $r_i$.

A value of 0 in this measure means that there are no words in the intersection, which indicates high level of variability in the results, while a value of 1 means that all the words are in the intersection, indicating that the results are fully consistent. We expect to see higher intersection@$k$ as $k$ grows. This expectation is confirmed by our experiments in Section~\ref{res:stability}.

We measure the stability of the usage-change detection algorithms with respect to a change in the underlying word embeddings: we apply the intersection@$k$ metric to two runs of the usage-change detection algorithm on the same corpus-pair, where each run is based on a different run of the underlying word embedding algorithm.

\section{The Predominant Approach}
\label{sec:align}

The most prominent method for detecting usage change is that of \newcite{HLJ16}, originally applied to detect shifts in dominant word senses across time. It is still the predominant approach in practice,\footnote{This is also indicated by the large number of citations: 350 according to Google Scholar.} with recent works building upon it \cite{YSD18,RB18}. This method was also shown to be the best performing one among several others \cite{SHD19}.

It works by training word embeddings on the two corpora, aligning the spaces, and then ranking the words by the cosine-distance between their representations in the two spaces, where large distance is expected to indicate significant change in meaning. We refer to this method as AlignCos.

The alignment is performed by finding an orthogonal linear transformation $Q$ that, when given matrices $X$ and $Y$, projects $X$ to $Y$ while minimizng the squared loss:

\[
Q = \arg\min_{Q}||QX-Y||_2, \hspace{0.3cm}  \text{s.t. $Q$ is orthogonal}
\]

The rows of $X$ correspond to embeddings of words in space A, while the rows of $Y$ are the corresponding embeddings in space B.
This optimization is solved using the Orthogonal Procrustes (OP) method \cite{S66}, that provides a closed form solution. 

Vector space alignment methods are extensively studied also outside of the area of detecting word change, primarily for aligning embedding spaces across language pairs \cite{XWL15,ALA17,LCD18,ALA18}. Also there, the Orthogonal Procrustes method is taken to be a top contender \cite{CLR17,KRC18}.

\subsection{Shortcomings of the alignment approach}

\paragraph{Self-contradicting objective.}
Note that the optimization procedure in the (linear) alignment stage attempts to project each word to itself. This includes words that changed usage, and which therefore should not be near each other in the space. While one may hope that other words and the linearity constraints will intervene, the method may succeed, by mistake, to project words that did change usage next to each other, at the expense of projecting words that did not change usage further apart than they should be. This is an inherent problem with any alignment based method that attempts to project the entire vocabulary onto itself.

\paragraph{Requires non-trivial filtering to work well.}
In addition, the alignment-based method requires non-trivial vocabulary filtering to work well. For example, \newcite{HLJ16} extensively filter proper nouns. Indeed, without such filtering, proper-nouns dominate the top of the changed words list. This does not indicate real word usage change, but is an artifact of names being hard to map across embedding spaces. In that respect, it makes sense to filter proper nouns. However, \emph{some} cases of word usage change do involve names. For example,  the word ``Harlem", which is used as either a name of a neighborhood in NY or as a name of a basketball team, was detected by our method as a word whose usage changed between tweets of celebrities with different occupations (\S\ref{res:detected}).

\paragraph{Not stable across runs.}
As we discuss in Section \ref{sec:stability} and show in Section \ref{res:stability}, the approach is not very stable with respect to different random seeds in the embeddings algorithm.

\begin{table*}[]
\scriptsize
\begin{center}
    \resizebox{\textwidth}{!}{
    \begin{tabular}{l||l|l|l|l|l|l|l|l|l|l|l|l|l}
             & \multicolumn{2}{c|}{Age} & \multicolumn{2}{c|}{Gender} & \multicolumn{3}{c|}{Occupation} & \multicolumn{2}{c|}{Day-of-week} & \multicolumn{2}{c|}{Hebrew} & \multicolumn{2}{c}{French} \\ \hline \hline
             & young       & older       & male        & female       & creator  & sports  & performer & weekday       & weekend  & 2014         & 2018        & 2014         & 2018             \\ \hline
    \#words  & 58M         & 116M      & 293M        & 126M         & 87M      & 132M    & 126M  & 142M          & 114M     & 42M          & 155M        & 867M         & 1B                  \\ \hline
    \#tweets & 5M          & 8M        & 23M         & 9M           & 6M       & 11M     & 10M    & 9M            & 7M     & 4M           & 13M         & 82M          & 104M                 \\ \hline
    \#vocab  & 42K         & 73K       & 114K        & 69K          & 63K      & 66K     & 69K    & 81K           & 72K   & 84K          & 187K        & 263K         & 350K         \\ \hline        
    \end{tabular}
    }
    \caption{Statistics of the different splits.}
    \label{tab:corpus_stats}
\end{center}
\end{table*}

\section{Nearest Neighbors as a Proxy for Meaning}
\label{sec:proposal}

Rather than attempting to project two embedding spaces into a shared space (which may not even map 1:1), we propose to work at the shared vocabulary space. The underlying intuition is that words whose usage changed are likely to be interchangeable with different sets of words, and thus to have different neighbors in the two embedding spaces.
This gives rise to a simple and effective algorithm:
we represent each word in a corpus as the set of its top $k$ nearest neighbors (NN). We then compute the score for word usage change across corpora by considering the size of the intersection of the two sets (not to be confused with intersection@$k$ defined in Section~\ref{sec:stability}): 

\begin{equation}
\mathrm{score}^k(w) = - | NN^k_1(w) \cap NN^k_2(w)|
\end{equation}
where $NN^k_i (w)$ is the set of k-nearest neighbors of word $w$ in space $i$.
Words with a smaller intersection are ranked higher as their meaning-change potential.

We only consider the words in the intersection of both vocabularies, as words that are rare in one of the corpora are easy to spot using the frequency in the two spaces, and do not neatly fit the definition of \textbf{usage} change.

Note that our method does not require extensive filtering of words -- we only filter words based on their frequency in the corpus\footnote{For English experiments we also filter stopwords according to the predefined list from NLTK.}.

We use a large value of $k=1000$\footnote{While this value may seem arbitrary, we tested several values in that range which yielded very similar results. However, the appropriate range may change when used with smaller corpora, or substantially different vocabulary sizes. We consider $k$ to be the only hyperparameter of our method, and note that it is rather easy to set.} in practice, because large neighbor sets are more stable than small ones \cite{WKM18}, leading to improved stability for our algorithm as well.

\paragraph{Limitations}
Similar to previous methods, our method assumes high quality embeddings, and hence also a relatively large corpus. Indeed, in many cases we can expect large quantities of data to be available to the user, especially when considering the fact that the data needed is raw text rather than labeled text. Using a limited amount of data results in lower quality embeddings, but also with smaller vocabulary size, which might affect our method. For high-quality embeddings with small vocabulary sizes, we believe that changing $k$ accordingly should suffice. Naturally, results will likely degrade as embeddings quality deteriorate.

It is also important to note that, like previous approaches, our method does not attempt to provide any guarantees that the detected words have indeed undergone usage change. It is only intended to propose and highlight \textbf{candidates} for such words. These candidates are meant to later be verified by a user who needs to interpret the results in light of their hypothesis and familiarity with the domain. Unlike previous methods, as we discuss in Section~\ref{sec:interpretation}, our method also provides intuitive means to aid in such an interpretation process.

\section{Experimental Setup}
\label{experimental}

We compare our proposed method (\textbf{NN}) to the method of \newcite{HLJ16} described in Section \ref{sec:align} (\textbf{AlignCos}), in which the vector spaces are first aligned using the OP algorithm, and then words are ranked according to the cosine-distance between the word representation in the two spaces.\footnote{Some extensions may yield improved results (filtering out proper names, as done in \newcite{HLJ16}, or jointly learning and aligning the spaces \cite{BM17,RRAB17,RB18,YSD18}, but we stick to this setting as it is the most general out of this line of work, and the one most commonly used in practice, for which an open implementation is available.} This method was shown to outperform all others that were compared to it by \newcite{SHD19}.

We demonstrate our approach by using it to detect change in word usage in different scenarios. We use the following corpora, whose statistics are listed in Table~\ref{tab:corpus_stats}.

We consider three demographics-based distinctions (age, gender, occupation), a day-of-week based distinction, and short-term (4y) diachronic distinctions.  We also compare to the longer-term (90y) diachronic setup of \newcite{HLJ16}, which is based on Google books.

\paragraph{Author Demographics} The Celebrity Profiling corpus~\cite{WSP19} 
consists of tweets from celebrities along with their traits such as age, gender and occupation. Based on these labels, we create the following splits: (1) \textbf{Age}: Young (birthyear 1990--2009) vs. Older (birthyear 1950--1969); (2) \textbf{Gender}: Male vs. Female; (3) \textbf{Occupation}: pairwise splits with Performer, Sports and Creator.

\paragraph{Day-of-week} \newcite{YL11} collect $580$ million tweets in English from June 2009 to February 2010, along with their time-stamps. As this is a fairly large corpus, we consider the tweets of a single month (November 2009). We create a split based on the Day-of-Week: weekday (tweets created on Tuesday and Wednesday) vs. weekend (tweets created on Saturday and Sunday). We remove duplicated tweets, as preliminary experiments revealed odd behavior of the representations due to heavily duplicated spam tweets. 

\paragraph{French Diachronic (4y, tweets)} \newcite{AKMCF18} compile a collection of tweets in French between the years 2014 and 2018. The authors utilize several heuristics based on the users' spatial information to consider tweets from users based in French territory only. We use the 2014 and 2018 portions of the data, and create a split accordingly. 

\paragraph{Hebrew Diachronic (4y, tweets)}

The Hebrew data we use is taken from a collection of Hebrew tweets we collected for several consecutive years, up to 2018. The collection was performed by using the streaming API and filtering for tweets containing at least one of the top 400 most frequent Hebrew words.
We use the 2014 and 2018 portions of the data, and create a split accordingly.

\paragraph{English Diachronic (90y, books)}
For diachronic study on English corpora, we make use of the embeddings trained on Fiction from Google Books~\cite{DAV15} provided by the authors of \newcite{HLJ16}, specifically for the two years, 1900 and 1990. These embeddings are originally aligned using Orthogonal Procrustes and the words whose relative frequencies are above $10^{-5}$ in both the time periods are ranked using cosine distance.

\subsection{Implementation details}
\label{training}

\paragraph{Tokenization and Word Embeddings} 

We use 300 dimensions word2vec vectors with 4 words context window. Further details of embeddings algorithm and tokenization are available in the appendix.

\paragraph{Vocabulary and Filtering}

We perform frequency-based filtering of the vocabulary, removing stop words (the most frequent 200 words for each corpus, as well as English stop words as defined in \texttt{\small nltk}\footnote{\url{https://www.nltk.org/}}), as well as low frequency words (we discard the 20\% least frequent words in each corpus, and require a minimum of 200 occurrences).

Notably, \textbf{we do not perform any other form of filtering}, and keep proper-nouns and person-names intact.

We consider neighbors having a raw frequency greater than 100 and identify 1000 such nearest neighbors ($k=$1000) to perform the intersection.

\begin{table*}[h!]
    
	\begin{center}
	    \resizebox{\textwidth}{!}{
		\begin{tabular} {c||l}
\hline
            \multicolumn{2}{l}{{\sc Age (Young vs. Older)}} \\\hline
            {\bf NN} & neighbors in each corpus \\ \hline \hline
            \multirow{2}{*}{dem} &  dese, yuh, them, nuh, dey, ayye, dats, tha, betta, fuk \\                   
            &  repub, democrats, centrist, manchin, primaries, party's, alp, dfl, gopers, repubs\\ \hline
            \multirow{2}{*}{dam} &  damm, mannnnn, mannnn, mane, huh, ahh, oo, buggin, koo, mannn \\                   &  dams, basin, river, dredging, reservoir, drainage, wastewater, sewerage, refinery, canal\\ \hline
            \multirow{2}{*}{rep} &  reppin, wear, allegiance, all-american, wildcat, alumni, tryout, hoosier, recruit, ua \\                   
            &  sen., congresswoman, chairwoman, co-chairs, gazelka, salazar, amb, comptroller, staffer, cong\\ \hline
            \multirow{2}{*}{assist} &  points, shutout, scoresheet, scored, pts, hatrick, sheet, nil, sacks, \includegraphics[scale=0.2]{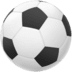} \\ &  assisting, contact, coordinate, locating, coordinating, administer, equip, consular, deploy, locate\\ \hline
            \multirow{2}{*}{pr} &  cameron, -pr, erik, lap, sargeant, laps, tundra, teamjev, caution, restart \\  &  stunt, puerto, promotional, rico, creative, ploy, hire, spin, freelance, fema\\ \hline
            \multirow{2}{*}{fr} &  frfr, forreal, foreal, lmaooo, madd, tho, bck, bruhh, lmao, fwm \\                   &  pavone, incl, from, wrk, ger, joseph, covey, env, w, ans\\ \hline
            \multirow{2}{*}{joint} &  jawn, fusion, scorpion, sumn, spot, db, cb, joints, mgmt, fye \\          
            &  high-level, convened, minsk, two-day, bilateral, counter-terrorism, delegations, asean, convene, liaison\\ \hline
            \multirow{2}{*}{mega} &  \includegraphics[scale=0.2]{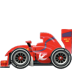}, fantastic, simulator, macau, lotus, fuji, bmw, awesome, mclaren, fab \\                   
            &  gujarat, becos, multi-billion, gta, rupees, dollar, maharashtra, major, crores, multi-million\\ \hline
            \multirow{2}{*}{flow} &  beard, vibin, jeezy, drizzy, lite, mohawk, dreads, sauna, boomin, vibe \\          &  illicit, influx, accumulation, moisture, absorb, overwhelm, heart's, drains, curtail, diverting\\ \hline
            \multirow{2}{*}{icymi} &  superintendent, bureau, commissioner, spokesman, exec, state's, prosecutor, reuters, montgomery, conway \\                  
            &  re-upping, reichert, newsmakers, sherrod, column, arizona's, otl, holcomb, rundown, wrap-up\\ \hline \hline
            {\bf AlignCos top-10} &  {\bf leo, whip, savage, nd, cole, pb, ace,
            carter, fr, bb}\\
\hline
		\end{tabular}}
		
              \caption{Top-10 detected words from our method (NN) vs.
                AlignCos method (last row), for corpus split according
                to the age of the tweet-author. Each word from our
                method is accompanied by its top-10 neighbors in each
                of the two corpora (Young vs. Older).}
		\label{tab:top10_age}
	\end{center}
	
\end{table*}

\section{Results}
\label{sec:experiments}

\subsection{Qualitative Evaluation: Detected Words}
\label{res:detected}
We run our proposed method and AlignCos \cite{HLJ16} on the different scenarios described in Section \ref{experimental}, and manually inspect the results. While somewhat subjective, we believe that the consistent success on a broad setting, much larger than explored in any earlier work, is convincing. We provide examples for two of the setups (English Diachronic and Performer vs. Sports), with the rest of the setups in the appendix. For each one, we list a few interesting words detected by the method, accompanied by a brief explanation (according to the neighbors in each corpus).

In addition, we depict the top-10 words our method
yields for the \textbf{Age} split (Table~\ref{tab:top10_age}), accompanied by the nearest neighbors in each corpus (excluding words in the
intersection), to better understand the context. For comparison, we
also mention the top-10 words according to the AlignCos method. Similar tables for the other splits are provided in the Appendix.

\paragraph{Across all splits,} our method is able to detect high quality words as
words that undergo usage change, most of them easily explained by
their neighboring words in the two corpora. As expected, we see that
the AlignCos method \cite{HLJ16} is highly sensitive to names, featuring many in the top-10 lists across the different splits. As opposed to AlignCos, our method is robust to global changes in the embedding space, since it looks at many neighbors. As a result, it is not sensitive to groups of words that ``move together'' in the embedding space (which might be the case with names).

\paragraph{English (diachronic, 90y)} Top-100 words identified by our method
cover all the words attested as real semantic shift in
\newcite{HLJ16}'s top-10 except the word `wanting'. Specifically,
three attested words, `gay', `major' and `check' are present in our
top-10, which also has more interesting words not present in
\newcite{HLJ16}'s top-10 (1900 vs. 1990): \textbf{van} (captain vs.
vehicle), \textbf{press} (printing vs.  places), \textbf{oxford}
(location vs. university). In addition, interesting words that came up
in the top-30 list are the following: \textbf{headed} (body part vs.
move in a direction), \textbf{mystery} (difficulty in understanding
vs. book
genre).

\paragraph{Occupation (performer vs. sports)}

Interesting words found at the top-10 list are the following: \textbf{cc} (carbon copy vs. country club), \textbf{duo} (duet vs. pair of people), \textbf{wing} (politics vs. football player position). In addition, interesting words that came up in the top-30 list are the following: \textbf{jazz} (music genre vs. basketball team), \textbf{worlds} (general meaning vs. championships), \textbf{stages} (platforms vs. company(bikes)), \textbf{record} (music record vs. achievement), \textbf{harlem} (neighborhood vs. basketball team).

\begin{figure*}[t!]
    \centering
    \begin{subfigure}[t]{0.32\textwidth}
        \centering
        \includegraphics[height=1.6in, width=2in]{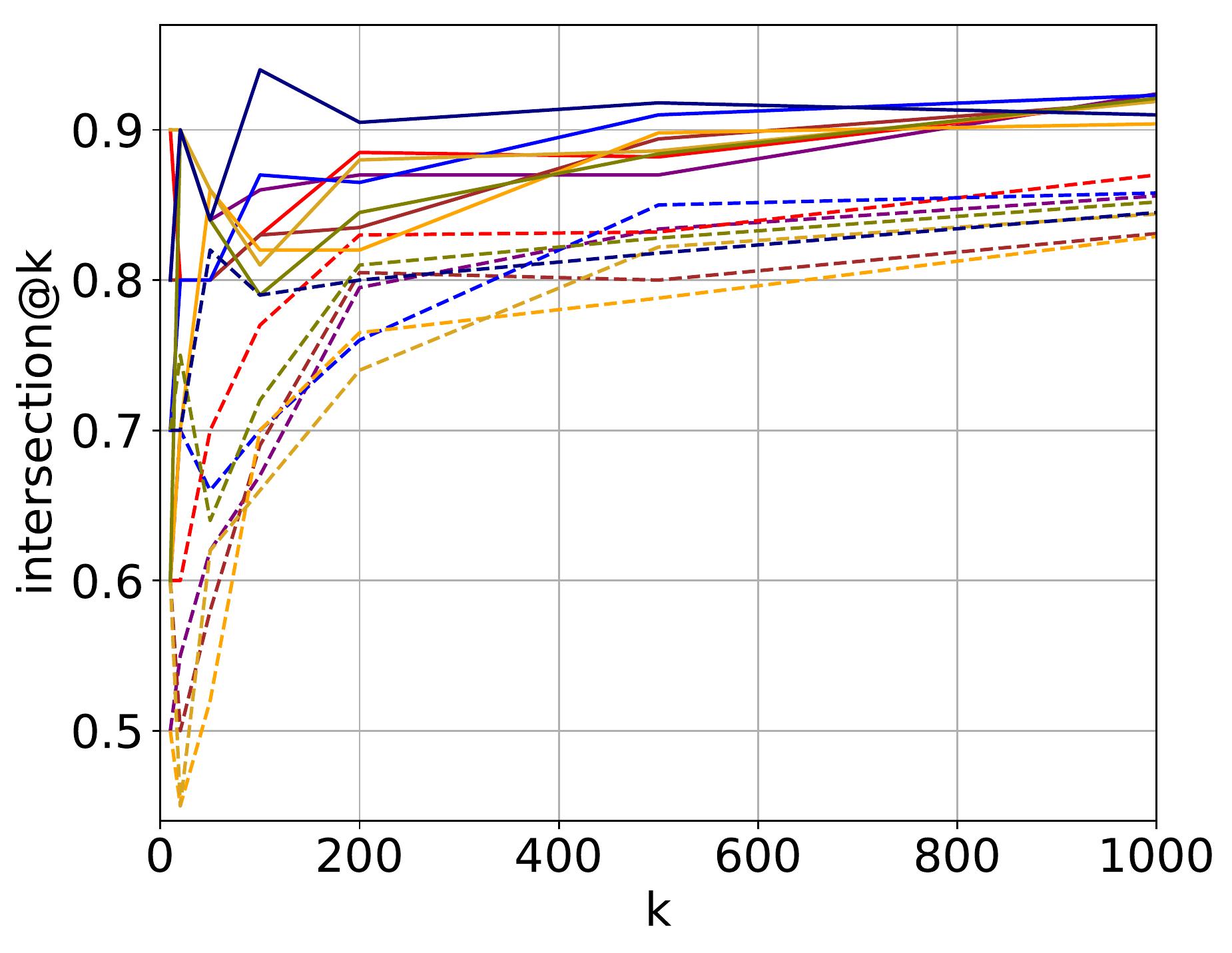}
        \caption{Change in intersection@$k$ w.r.t $k$. }

    \end{subfigure}
    \begin{subfigure}[t]{0.30\textwidth}
        \centering
        \includegraphics[height=1.6in, width=2in]{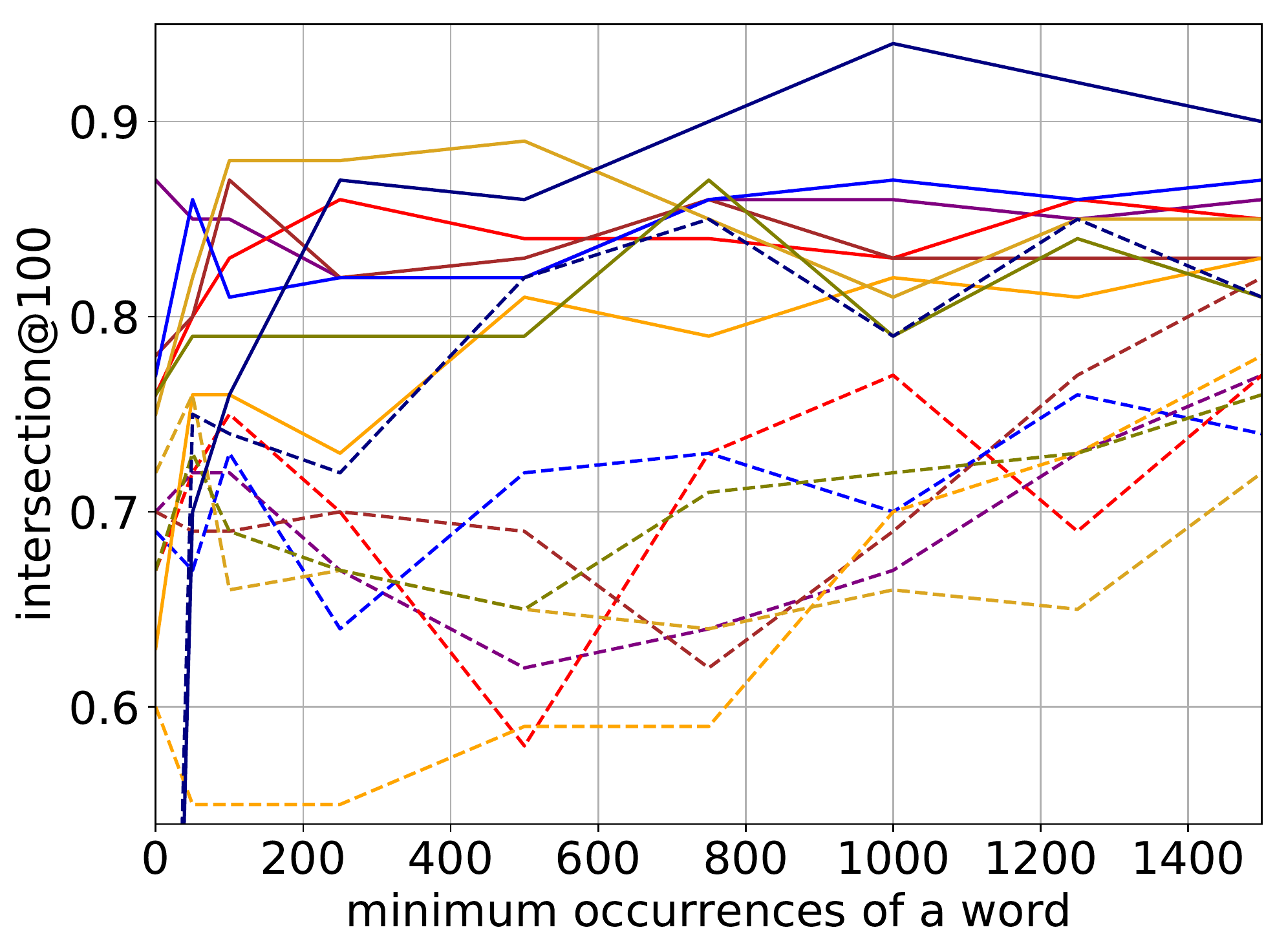}
        \caption{Change in intersection@100 w.r.t word frequency cut-off.}
    \end{subfigure}
    \hspace{1.8mm}
    \begin{subfigure}[t]{0.32\textwidth}
        \centering
        \includegraphics[height=1.6in, width=2in]{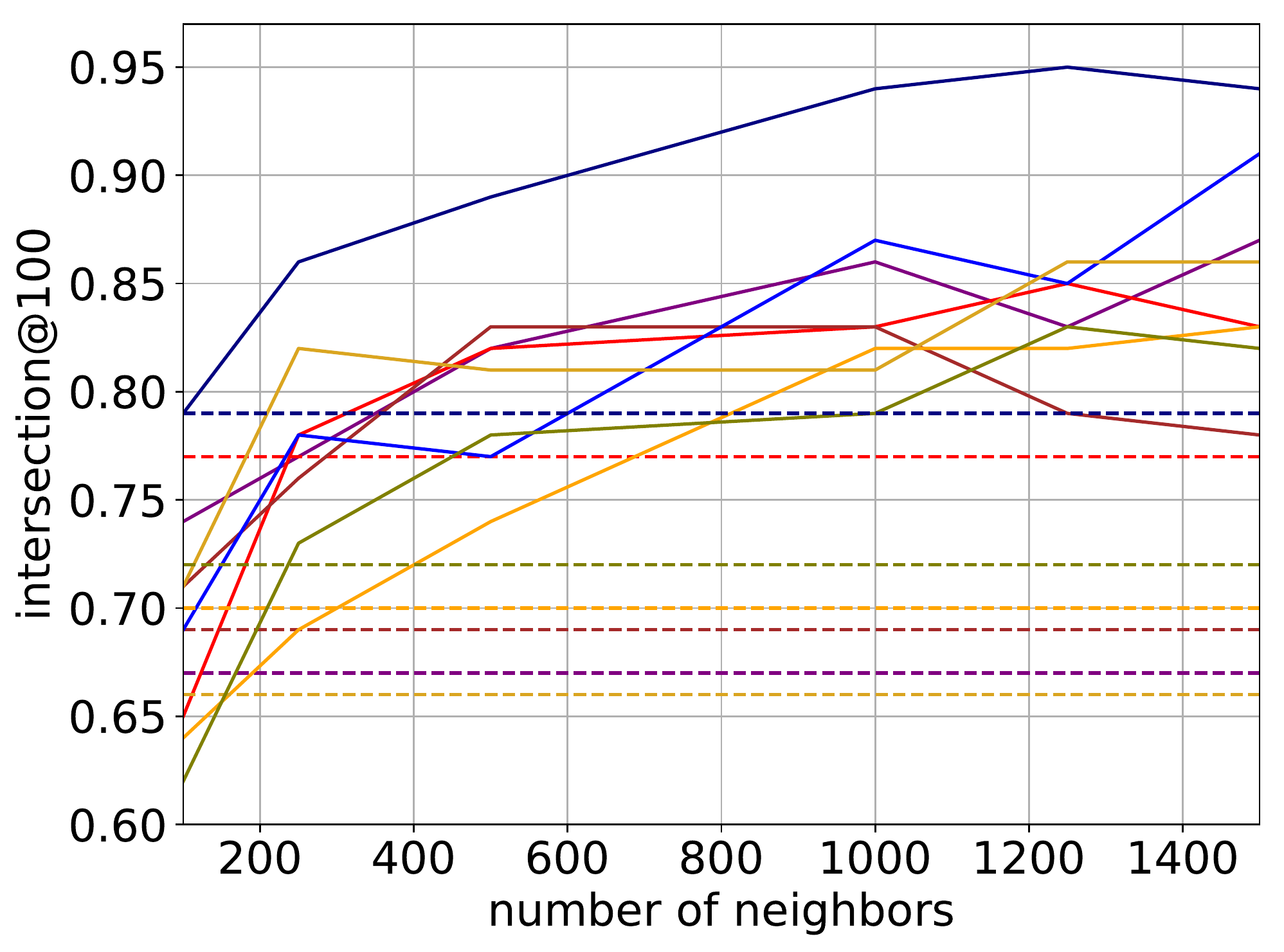}
        \caption{Change in intersection@100 w.r.t number of neighbors to consider.}
    \end{subfigure}
    \begin{subfigure}[t]{1\textwidth}
        \centering
        \includegraphics[height=0.6in,width=5.5in]{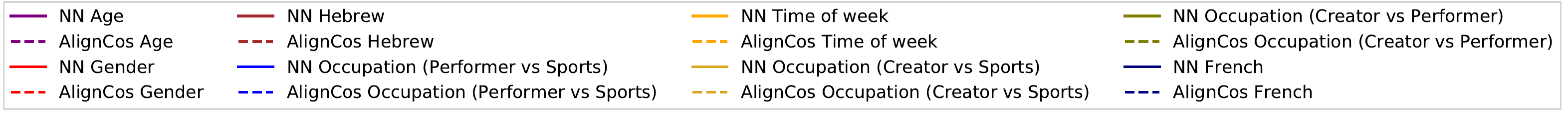}
    \end{subfigure}
    \caption{Stability plots.  Solid lines: our method, dashed lines: AlignCos method.}
    \label{fig:stability_neighbors}
    
\end{figure*}

\subsection{Quantitative Evaluation: Stability}
\label{res:stability}
We compare the stability of our method to that of the AlignCos method \cite{HLJ16} using the intersection@$k$ metric, as defined in Section~\ref{sec:stability}. We use $k \in {10, 20, 50, 100, 200, 500, 1000}$.

In Figure~\ref{fig:stability_neighbors}(a) we plot the intersection@$k$ for different values of $k$ for all splits, with solid lines for the results of our method and dashed lines for the results of AlignCos method. It is clear that our method is significantly more stable, for all $k$ values and across all splits. To better understand the parameters that affect the stability of the different methods, we also examine how the intersection changes with different values of frequency cut-off. In Figure~\ref{fig:stability_neighbors}(b) we plot intersection@100 as a function of the frequency cut-off (minimum word occurrences required for a word to be included in the ranking). Here, our method is again more stable for all corpus splits. In addition, our method is similarly stable, regardless the frequency cut-off, unlike the AlignCos method. We also examine how the size of NN lists considered for the intersection affects the stability. In Figure~\ref{fig:stability_neighbors}(c) we plot the intersection@100 against number of neighbors taken into consideration using our method. We get that from around $k=250$, our method is substantially more stable for all splits.

\begin{figure*}[t!]
    \centering
    \begin{subfigure}[t]{0.48\textwidth}
        \centering
        \includegraphics[height=2in, width=2.7in]{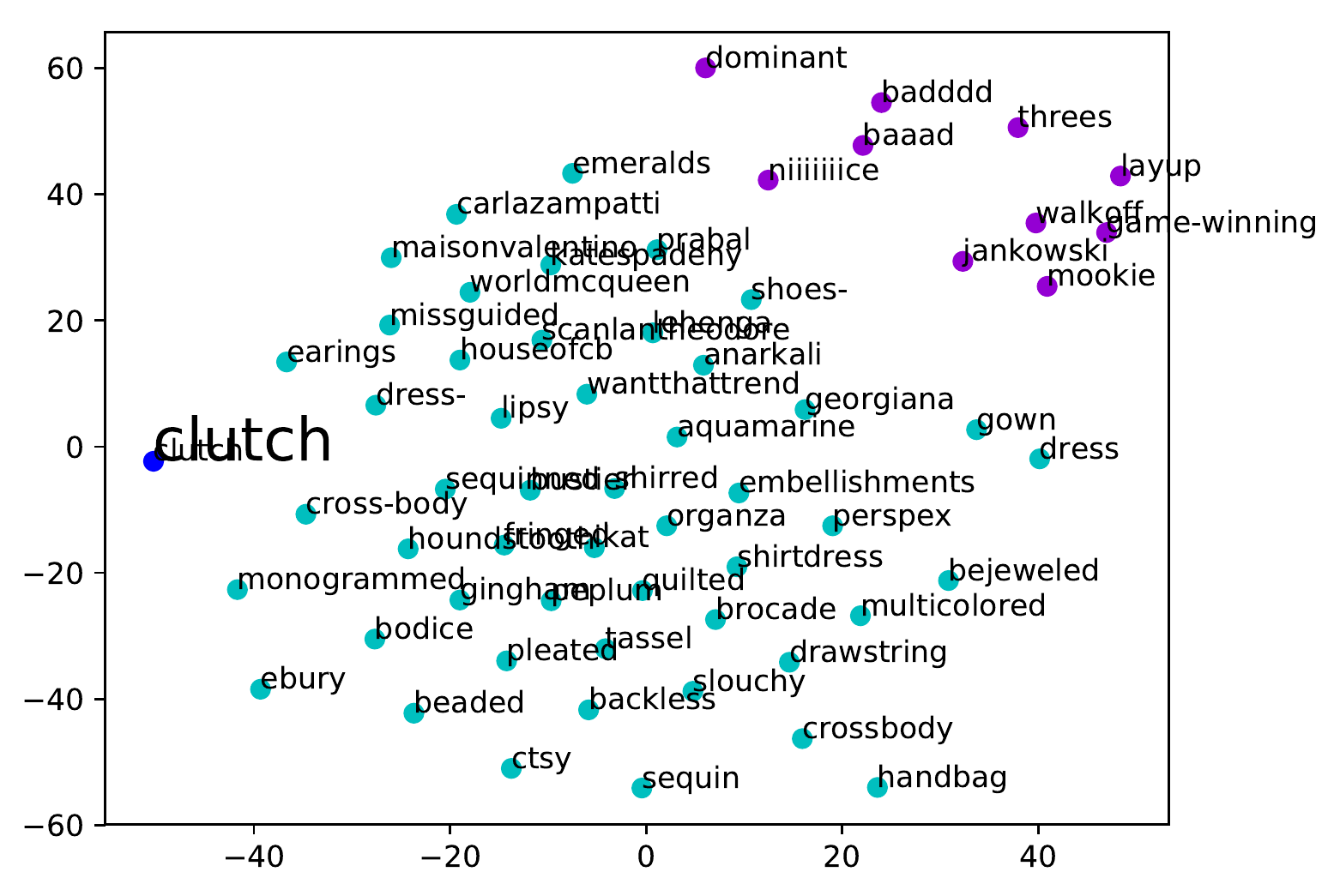}
        \caption{Female space}
    \end{subfigure}
    \begin{subfigure}[t]{0.48\textwidth}
        \centering
        \includegraphics[height=2in, width=2.7in]{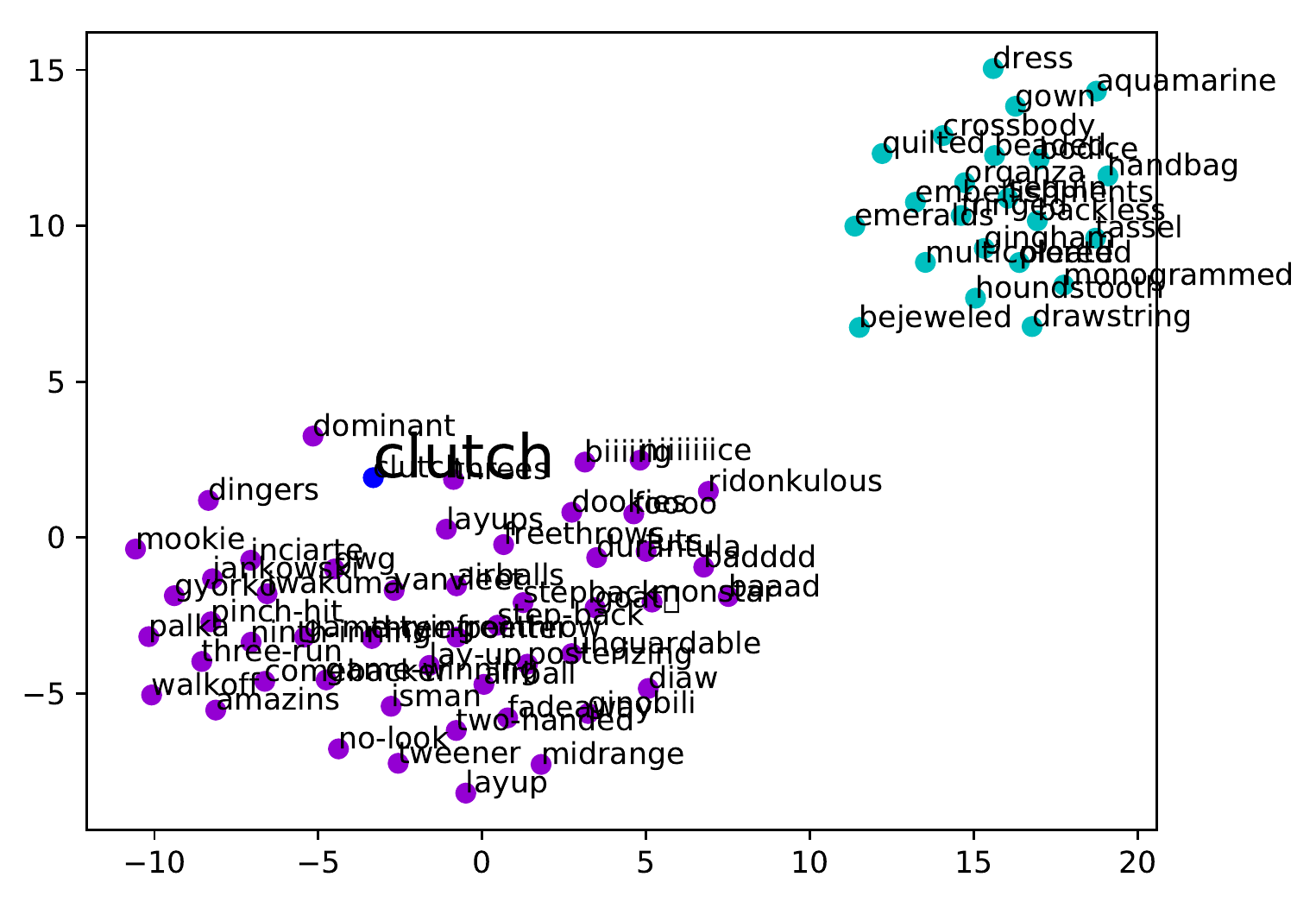}
        \caption{Male space}
    \end{subfigure}
    \caption{t-SNE visualization of top-50 neighbors from each corpus for word `clutch', Gender split, with cyan for female and violet for male.}
    \label{fig:vis_gender_clutch}
\end{figure*}

\begin{figure*}[t!]
    \centering
    \begin{subfigure}[t]{0.48\textwidth}
        \centering
        \includegraphics[height=2in, width=2.7in]{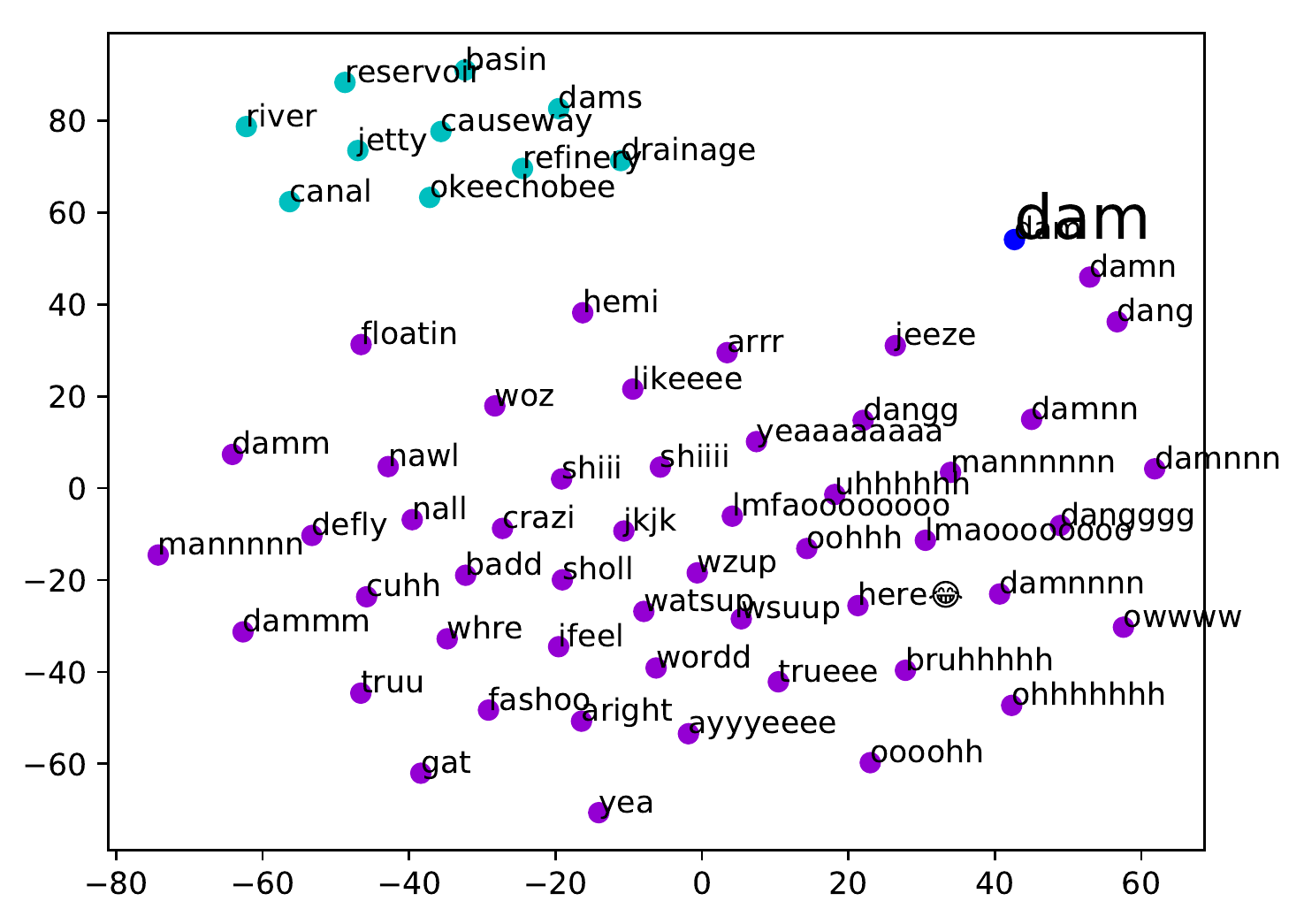}
        \caption{Young space}
    \end{subfigure}
    \begin{subfigure}[t]{0.48\textwidth}
        \centering
        \includegraphics[height=2in, width=2.7in]{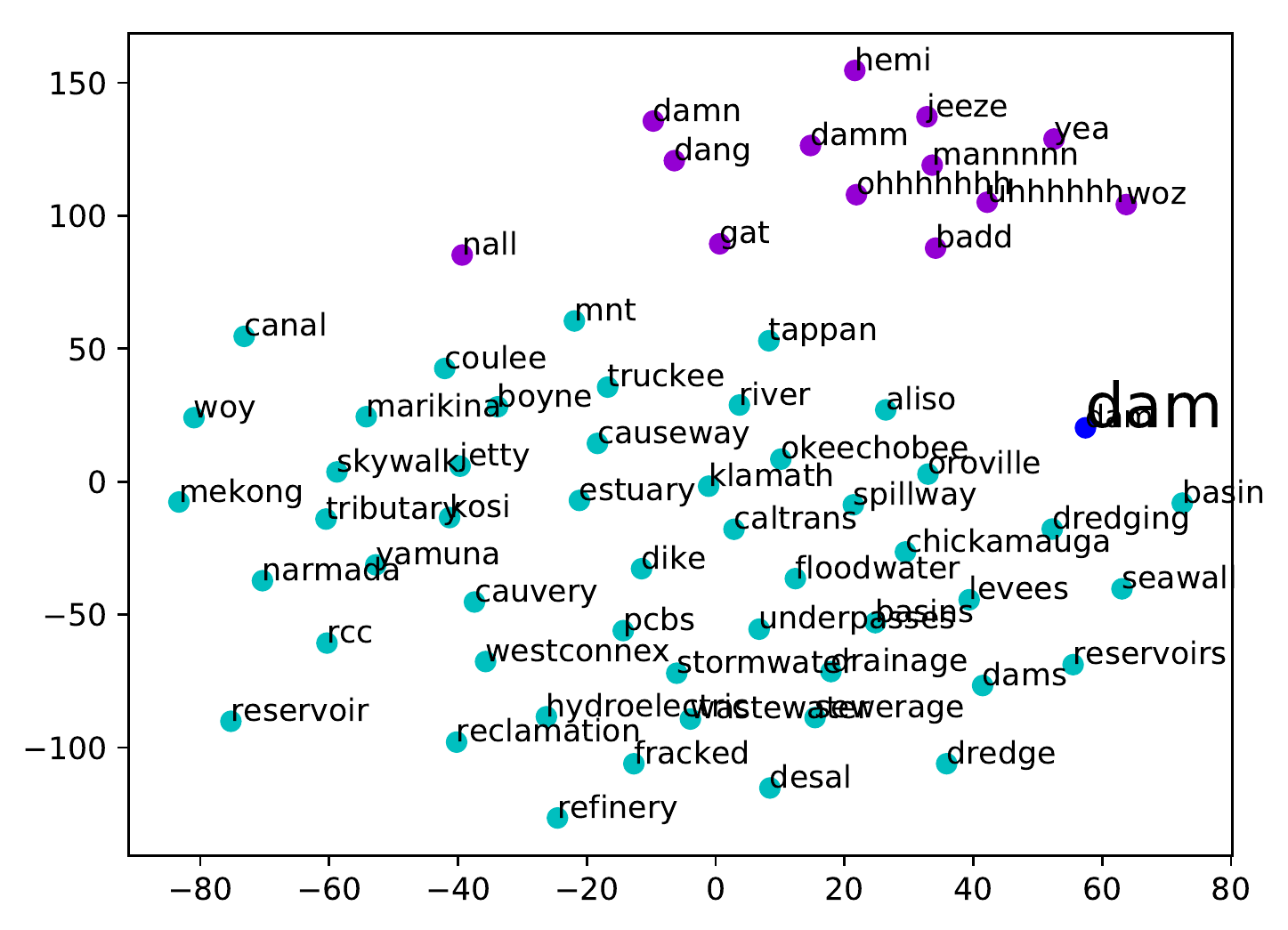}
        \caption{Older space}
    \end{subfigure}
    \caption{t-SNE visualization of top-50 neighbors from each corpus for word `dam', Age split, with cyan for older and violet for young.}
    \label{fig:vis_age_dam}
\end{figure*}

\subsection{Quantitative Evaluation: DURel and SURel datasets}
\label{res:wind}

This field of semantic change suffers from lack of proper evaluation datasets, and there is no common benchmark that is being used. Two new datasets were recently introduced, and used to extensively compare between previous methods \cite{SHD19}: the DURel dataset \cite{SSE18} focuses on diachronic changes, while the SURel dataset \cite{HSS19} focuses on domain-based semantic changes. We use them to verify the quality of our results and compare against AlignCos \cite{HLJ16}.

Both datasets include a limited number of German words, along with human annotations of the degrees of semantic relatedness between contexts of the words (across the different texts). However, they are not ideal as they are extremely limited (22 words each)\footnote{For our experiments, we follow the setup of \newcite{SHD19} and use 19/21 words for DURel/ SURel respectively.}.

\paragraph{Evaluation Metrics} 
Spearman correlation is the standard measure used in this field to compare between methods with respect to gold rankings. However, it is extremely important to note its limitations in this setting, since comparing to a very small gold ranking might be tricky. Specifically, it does not take into account the global ranking of each method, but only the relative position of each of the gold words in each method's ranking. For example, a method that ranks all the gold words at the bottom of the ranking (out of all the words in the vocabulary) in the same order, would be considered perfect, even though it is clearly not the case.

As a possible solution for this problem, we suggest to use Discounted Cumulative Gain (DCG), which better captures also global rankings. As opposed to Spearman, this measure takes into account not only the order of the words, but also their actual scores:
\begin{equation}
\mathrm{DCG(M)} = \sum_{w \in W}{\frac{GoldScore(w)}{\log_2(rank_M(w) + 1)}}
\end{equation}
where $W$ are the words in the gold dataset, and $M$ is the model being evaluated.

\begin{table}[]

\centering
\begin{tabular}{l|l|l|l}
\hline
method   & measure  & SURel & DURel \\ \hline
AlignCos  & spearman & 0.800   & 0.814  \\
NN       & spearman & 0.859 & 0.59  \\
AlignCos & DCG      & -4.5  & -4.31 \\
NN       & DCG      & -4.54 & -4.3 \\ \hline
\end{tabular}
\caption{Results on DURel and SURel with NN and with AlignCos.}
\label{tab:durelsurel}
\end{table}

We report the results in Table~\ref{tab:durelsurel}. We compute AlignCos results with the best parameters reported in \newcite{SHD19}\footnote{We were unable to reproduce the exact results from the paper: spearman correlation of 0.866 and 0.851 on SURel and DURel, respectively.}. Our method outperforms AlignCos on SURel, both when measuring with spearman correlation\footnote{Average Spearman score over model runs with different numbers of iterations, as done in \cite{SHD19}.} and with DCG. For DURel, AlignCos gets better results when measuring with spearman, but both methods are on par when using DCG.

\subsection{Interpretation and Visualization}
\label{sec:interpretation}

We find that in many cases, it is not clear why the returned candidate words were chosen, and questions such as ``why is the word `dam' different across age groups?'' often arise. 
The NN method lends itself to interpretation, by considering the top-10 neighbors, as shown in Table~\ref{tab:top10_age}. 
We note that this interpretation approach is very reliable in our method, as we are guaranteed to gain insights about the usage change when looking at neighboring words, since most of the neighbors will be different for the identified words. While we can definitely attempt at looking at the NN also for the OP-based methods, there we are not guaranteed at all to even spot a difference between the neighbors: it may absolutely be the case that the identified word moved in the embedding space “together” with most of its neighbors. In this case, looking at the neighbors will provide no insight on the nature of this change. We observed this phenomenon in practice. Nonetheless, comparing flat word lists is hard, and 10 words are often insufficient.

We present a visualization method that aids in understanding the model's suggestions. The visualization consists of projecting the word of interest and its top-50 neighbors from each corpus into two dimensions using t-SNE \cite{MH08}, and plotting the result while coloring the neighbors in the intersection in one color and the neighbors unique to each corpus in other colors. We expect the neighbors of a word of interest to have distinct neighbors across the corpora.

Figures~\ref{fig:vis_gender_clutch} and~\ref{fig:vis_age_dam} show the visualizations for the word \textit{clutch} in the \textbf{Gender} split, with cyan for female and violet for male, and the word \textit{dam} in the \textbf{Age} split, with cyan for older and violet for young (in both cases they were no shared neighbours). We plot the projection of the words twice -- one plot for each embedding space. We can see that, as expected, the neighboring words are distinct, and that the target word belongs to the respective neighborhood in each space. We conclude that this is a useful tool for interpreting the results of our model.

\section{Related Work}

Extensive work has been done on detecting word usage change across corpora that predated the alignment-based methods \cite{MMR14,JD14,KWH15,HLS16,FL16}. 

In addition, two works are more closely related to our approach.
In \newcite{ADB17}, the authors also use the neighbors of a word in order to determine its stability (and therefore, the extent to which it changes). Their best model combines the traditional alignment-based approach with weighting the neighbors according to their rank and their stability. The algorithm is iterative, and they update the stability of all the words in the vocabulary in each update step. Our method uses the neighbors of the words directly, does not include an iterative process, and does not rely on cosine-distance in the aligned embeddings. In addition, their method requires computation for the whole vocabulary, while other methods, including ours, usually allow querying for a single word.

Another work that considers the neighbors of the word in order to determine the extent of change is that of \newcite{HLJ16b}, in which they suggest a measure that is based on the changes of similarities between the target word and its neighbors in both spaces. They find that this method is more suitable for identifying changes that are due to cultural factors, rather than linguistic shift. This may serve as another motivation to move from the global measures to a local one.

Two other relevant works have come to our attention after publishing this work, both using some variation of our method as a part of a different analysis, where identifying word usage change is the mean and not the end goal. Both have a very limited experimental setting compared to ours \cite{XK12,KK15}.

In a concurrent work, \newcite{RR19} model the
evolution of language in relation to world events. Their work deals with creating timelines of words based on semantic change and suggests, among other methods, to use the intersection of neighbours to detect relevant time points.

Recent works \cite{Giullianelli:2019:thesis,martinc-et-al:2020:lrec}
explored the possibility of modeling diachronic and usage change using
contextualized embeddings extracted from now ubiquitous Bert
representations \cite{devlin2019bert}. Focusing on the financial
domain, \newcite{montariol:2020:bertDiacChange} use, on top of Bert
embeddings, a clustering method that does not need to predefine the
number of clusters and which leads to interesting results on that
domain. Another approach from \newcite{hu-etal-2019-diachronic} relies
on the inclusion of example-based word sense inventories over time
from the Oxford dictionary to a Bert model. Doing so provides an
efficient fine-grained word sense representation and enables a
seemingly accurate way to monitor word sense change over time.  Most
of those approaches could be easily used with our method, the
inclusion of contextualized embeddings would be for example
straightforward, we leave it for future work.

\section{Conclusion}
Detecting words that are used differently in different corpora is an important use-case in corpus-based research. We present a simple and effective method for this task, demonstrating its applicability in multiple different settings. We show that the method is considerably more stable than the popular alignment-based method popularized by \newcite{HLJ16}, and requires less tuning and word filtering. We suggest researchers to adopt this method, and provide an accompanying software toolkit.

\section*{Acknowledgments}
We thank Marianna Apidianiaki for her insightful comments on an earlier version of this work. This project has received funding from the European Research Council (ERC) under the European Union's Horizon 2020 research and innovation programme, grant agreement No. 802774 (iEXTRACT), and from the the Israeli ministry of Science, Technology and Space through the Israeli-French Maimonide Cooperation programme. The second and third authors were partially funded by the French Research Agency projects ParSiTi (ANR-16-CE33-0021), SoSweet (ANR15-CE38-0011-01) and by the French Ministry of Industry and Ministry of Foreign Affairs via the PHC Maimonide France-Israel cooperation programme.

\bibliography{semantic_change}
\bibliographystyle{acl_natbib}

\clearpage

\appendix

\section{Implementation Details}

\paragraph{Tokenization} We tokenize the English, French and Hebrew tweets using ark-twokenize-py\footnote{\url{https://github.com/myleott/ark-twokenize-py}}, Moses tokenizer\footnote{\url{https://www.nltk.org/_modules/nltk/tokenize/moses.html}} and UDPipe~\cite{SS2017}, respectively. We lowercase all the tweets and remove hashtags, mentions, retweets and URLs. We replace all the occurrences of numbers with a special token. We discard all words that do not contain one of the following: (1) a character from the respective language; (2) one of these punctuations: ``-", ``'", ``."; (3) emoji.

\paragraph{Word embeddings}
We construct the word representations by using the continuous skip-gram negative sampling model from Word2vec~\cite{MCC13,MSC13}. We use the Gensim\footnote{\url{https://radimrehurek.com/gensim/models/word2vec.html}} implementation. For all our experiments, we set vector dimension to 300, window size to 4, and minimum number of occurrences of a word to 20. The rest of the hyperparameters are set to their default value.

For the \textbf{stability} experiments we run the embedding algorithm twice, each time with a different random seed.

\section{Qualitative Evaluation: Detected Words}

We show the top-10 words our method yields for each of the different splits, accompanied with the nearest neighbors in each corpus (excluding words in the intersection), to better understand the context. For comparison, we also show the top-10 words according to the AlignCos method. The splits are the following:

\paragraph{English: 1900 vs. 1990}
The list of top-10 detected words from our method (NN) vs. AlignCos method, for corpus split according to the year of the English text is displayed in Table~\ref{tab:hamilton_top10}.

\paragraph{Age: Young vs. Older}
The list of top-10 detected words from our method (NN) vs. AlignCos method, for corpus split according to the age of the tweet-author is displayed in Section \ref{sec:experiments}.
Interesting words found at the top-10 list are the following (young vs. older): \textbf{dem} (`them' vs. US political party), \textbf{dam} (`damn' vs. water barrier), \textbf{assist} (football contribution vs. help). In addition, interesting words that came up in the top-30 list are the following: \textbf{pc} (personal computer vs. Canadian party), \textbf{presents} (introduces vs. gifts), \textbf{wing} (general vs. political meaning), \textbf{prime} (general vs. political meaning), \textbf{lab} (school vs. professional).

\paragraph{Gender: Male vs. Female}
The list of top-10 detected words from our method (NN) vs. AlignCos method, for corpus split according to the gender of the tweet-author is displayed in Table~\ref{tab:gender_top10}. Interesting words found at the top-10 list are the following (male vs. female): \textbf{clutch} (grasping vs. female bag), \textbf{bra} (colloquial usage like `bro' vs. female clothing), \textbf{gp} (grand prix event vs. general practitioner). In addition, interesting words that came up in the top-40 list are the following: \textbf{stat} (statistics vs. right away), \textbf{pit} (car-related vs. dog-related), \textbf{dash} (radio station vs. quantity), \textbf{pearl} (pearl harbor vs. gemstone and color).

\paragraph{Occupation: Performer vs. Sports}
The list of top-10 detected words from our method (NN) vs. AlignCos method, for corpus split according to the occupation (Performer vs. Sports) of the tweet-author is displayed in Table~\ref{tab:occ_perf_sp_top10}.

\paragraph{Occupation: Creator vs. Sports}
The list of top-10 detected words from our method (NN) vs. AlignCos method, for corpus split according to the occupation (Creator vs. Sports) of the tweet-author is displayed in Table~\ref{tab:occ_cr_sp_top10}. Interesting words found at the top-10 list are the following
  (creator vs. sports): \textbf{cc} (carbon copy vs. country club),
  \textbf{op} (event opening vs. operation), \textbf{wing} (politics
  vs. football player position), \textbf{worlds} (earth vs. world
  cup).
In addition, interesting words that came up in the top-20 list
  are the following: \textbf{oval} (oval office vs. sports ground),
  \textbf{fantasy} (genre vs. fantasy football), \textbf{striking}
  (shocking vs. salient), \textbf{chilling} (frightening vs.
  relaxing), \textbf{fury} (book: fire and fury vs. British boxer).

\paragraph{Occupation: Creator vs. Performer}
The list of top-10 detected words from our method (NN) vs. AlignCos method, for corpus split according to the occupation (Creator vs. Performer) of the tweet-author is displayed in Table~\ref{tab:occ_cr_pe_top10}. Interesting words found at the top-10 list are the following (creator vs. performer): \textbf{dash} (travel vs. person), \textbf{presents} (introduces vs. gifts), \textbf{chapter} (book vs. movie). In addition, interesting words that came up in the top-30 list are the following: \textbf{cartoon} (cartoonist vs. movie), \textbf{scoop} (news story vs. ice cream), \textbf{mega} (money vs. largeness), \textbf{sessions} (assembly vs. period).

\paragraph{Time of week: Weekday vs. Weekend}
The list of top-10 detected words from our method (NN) vs. AlignCos method, for corpus split according to the time of week (Weekday vs. Weekend) of the tweet is displayed in Table~\ref{tab:time_of_week_top10}. Interesting words found at the top-10 list are the following (weekday vs. weekend): \textbf{cc} (credit card vs. carbon copy), \textbf{pitch} (presentation attribute vs. playing surface), \textbf{bond} (agreement vs. movie character). In addition, interesting words that came up in the top-30 list are the following: \textbf{sunday} (day of the week vs. vacation-related), \textbf{vp} (vice president vs. tv-series: True Jackson, VP), \textbf{third} (report-related vs. sports-related), \textbf{cliff} (first name vs. mountain cliff), \textbf{fight} (general meaning vs. boxing).

\paragraph{French: 2014 vs. 2018}
The list of top-10 detected words from our method (NN) vs. AlignCos method, for corpus split according to the year of the French text is displayed in Table~\ref{tab:fr_top10}. Interesting words found at the top-10 list are the following (2014 vs. 2018):  \textbf{ia} (frequent misspelled contraction of ``ya'' in 2014, vernacular form of {``il y a''}, \emph{there is}, vs. {``intelligence artificielle''}, \emph{artificial intelligence}),  \textbf{divergent} (the movie vs. the adjective). 
In addition, interesting words that came up in the top-30 list are the following: \textbf{pls} (contraction of the borrowing ``please'' vs. the acronym of ``Position lat\'erale de s\'ecurit\'e'', \emph{lateral safety position}, which is now used as a figurative synonym for ``having a stroke''. In the same vein, and tied to political debates, we note \textbf{apl} (contraction of ``appel/appeler'', \emph{call/to call}  vs. controversial housing subsidies).

\begin{figure*}
	\centering
	\includegraphics[width=15cm,height=10cm,keepaspectratio]{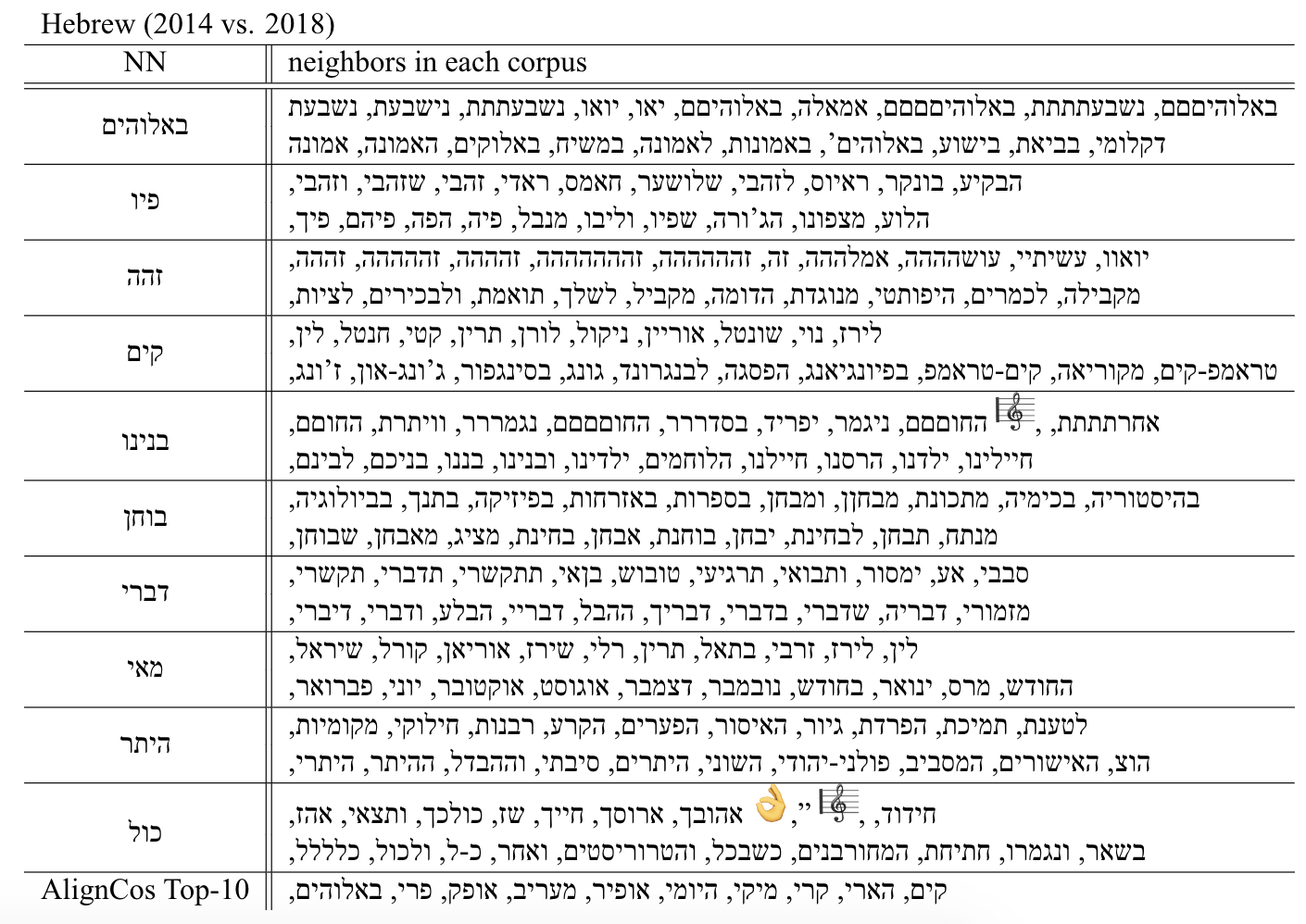}
	\caption{Top-10 detected words from our method (NN) vs. AlignCos method (last row), for corpus split according to the year of the text. Each word from our method is accompanied by its top-10 neighbors in each of the two corpora (2014 vs. 2018).}
	\label{fig:hebrew_top10}
\end{figure*}

\paragraph{Hebrew: 2014 vs. 2018}
The list of top-10 detected words from our method (NN) vs. AlignCos method, for corpus split according to the year of the Hebrew text is displayed in Figure~\ref{fig:hebrew_top10}. Interesting words found at the top-10 list (2014 vs. 2018) are the following (we use transliteration accompanied with a literal translation to English): \textbf{beelohim--\textit{in god}} (pledge word vs. religion-related) and \textbf{Kim--\textit{Kim}} (First name vs. Kim Jong-un). In addition, interesting words that came up in the top-30 list are the following: \textbf{shtifat--\textit{washing}} (plumbing vs. brainwashing), \textbf{miklat--\textit{shelter}} (building vs. asylum (for refugees)), \textbf{borot--\textit{pit/ignorance}} (plural of pit vs. ignorance).

\begin{table*}
    
	\begin{center}
	    \scalebox{0.75}{
		\begin{tabular} {c||l}

            \multicolumn{2}{l}{English (1900 vs. 1990)} \\\hline
            NN & neighbors in each corpus \\ \hline \hline
            \multirow{2}{*}{gay} &  cheery, humoured, apparel, natured, dresses, attire, neat, bright, genial, unusually \\                   
                                 &    lesbian, transgender, lesbians, katz, bisexual, bisexuals, coalition, gays, bi, gras \\ \hline
            \multirow{2}{*}{van} &  wyk, commented, sterne, skipper, south, simon, defarge, ned, island, carolina \\                   &   truck, helsing, luyden, luydens, pickup, toyota, jeep, porsche, volvo, der\\ \hline
            \multirow{2}{*}{press} &  pressed, publisher, papers, issues, dublin, circulation, thickest, wilson, paper, payment \\                   &  ams, belknap, harvester, wesleyan, newberry, westview, middletown, esp, harrington, gainesville\\ \hline
            \multirow{2}{*}{oxford} &  durham, albany, lincoln, sometime, ireland, john, canon, christ, bishops, newcastle \\                   &  clarendon, basingstoke, supervising, blackwell, 1921, researching, database, ibadan, walton, peruse\\ \hline
            \multirow{2}{*}{major} &  curtly, osborne, gordon, retorted, dryly, inspector, steele, chester, stewart, morris \\                   &  brigadier, factor, dramatist, producers, andre, schomburg, boswell, brian, biggest, insignia\\ \hline
            \multirow{2}{*}{2} &  vide, woodcuts, illustrations, peggy, demy, cloister, portrait, memoirs, baroness, allen \\                   &  rte, 767, tn, dresden, vols, 38225, bp, klingon, 1863, 98765432\\ \hline
            \multirow{2}{*}{cambridge} &  dublin, queens, glasgow, tutor, jesus, newcastle, christ, assistant, student, kent \\                   &  belknap, blackwell, 1921, persephone, harvester, hogarth, clarendon, ams, vols, esp\\ \hline
            \multirow{2}{*}{1} &  ornamental, woodcuts, dad, biography, section, demy, cent, 8vo, t, 3s \\                   & xlibris, deduct, freepost, 345, 1001, 98765432, 350, 888, toulouse, bunkyo\\ \hline
            \multirow{2}{*}{new} &  revised, comer, institute, commonwealth, comers, development, insurance, illustrated, testament, magazine \\                   &  ungar, picayune, schocken, ams, crowell, atheneum, upstate, 10012, praeger, harrington\\ \hline
            \multirow{2}{*}{check} &  restrain, effort, balance, exertion, strove, readiness, restrained, gave, jerk, held \\                   & cashier, update, checkbook, checks, payable, money, certificate, postal, brochure, lor\\ \hline

            AlignCos Top-10 & wanting, gay, check, starting, major, actually, touching, harry, headed, romance 
		\end{tabular}}
		
		\caption{Top-10 detected words from our method (NN) vs. AlignCos method (last row), for corpus split according to the year of the text. Each word from our method is accompanied by its top-10 neighbors in each of the two corpora (1900 vs. 1990).} 
		\label{tab:hamilton_top10}
	\end{center}
	
\end{table*}

\begin{table*}
    
	\begin{center}
	    \scalebox{0.85}{
		\begin{tabular} {c||l}

            \multicolumn{2}{l}{Gender (male vs. female)} \\\hline
            \multirow{2}{*}{bra} &  bruh, brah, bro, cuh, homie, boi, cuzzo, dawg, breh, brudda \\                  
            &  thong, jeans, strapless, leggings, tights, underwear, skirt, pants, sneakers, shorts\\ \hline
            \multirow{2}{*}{clutch} &  threes, walkoff, mookie, dingers, layups, midrange, game-winning, diaw, gwg, layup \\                   
            &  sequin, beaded, gown, dress, handbag, chiffon, headpiece, tote, sandal, swarovski\\ \hline
            \multirow{2}{*}{mm} &  cores, thickness, oled, diameter, deg, usb-c, ssd, dbo, gpu, cpu \\           &  \includegraphics[scale=0.2]{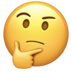}, huh, arizona, that's, errrr, bcz, thts, cc, \includegraphics[scale=0.2]{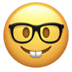}, \includegraphics[scale=0.2]{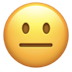}\\ \hline
            \multirow{2}{*}{mc} &  armand, dilla, rza, kuntryking, rapper, boney, riz, donald's, huss, dizzee \\        
            &  obe, showstopper, groupie, fleming, thnks, hoff, cohost, honoree, harmon, reece\\ \hline
            \multirow{2}{*}{gp} &  motogp, thruxton, monza, indycar, dtm, snetterton, suzuka, hockenheim, criterium, wec \\              
            &  physicians, pharmacists, clinical, procurement, ndis, insurers, nbn, tfl, hep, mh\\ \hline
            \multirow{2}{*}{keeper} &  midfielder, cech, krul, benteke, free-kick, freekick, aguero, defoe, benzema, goalscorer \\                 &  dynamo, goofball, hero, hustler, touche, stud, digger, nemesis, saver, ruler\\ \hline
            \multirow{2}{*}{nd} &  tht, iu, wvu, gtown, isu, wisco, ou, gng, huggs, byu \\         
            &  minot, nh, ky, hoosier, farmers, heitkamp, ranchers, dakota, rural, ndans\\ \hline
            \multirow{2}{*}{hay} &  bales, doon, beech, hinton, blackwood, noches, ayer, mong, dartford, rooty \\   
            &  beccy, goat, mclaren, portage, ale, glasto, grafton, daffodils, cornish, crap\\ \hline
            \multirow{2}{*}{steph} &  lebron, kyrie, klay, harden, draymond, rondo, melo, delly, dwade, korver \\       
            &  chels, rach, leah, sam, liz, dani, trish, lovie, cait, kel\\ \hline
            \multirow{2}{*}{echo} &  homepod, orc, cortana, npc, oculus, undead, redstone, forked, emergent, echoed \\ &  paradiso, avalon, asbury, hyde, sondheim, colosseum, oasis, \includegraphics[scale=0.2]{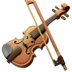}, empress, inconvenient\\ \hline \hline
            AlignCos Top-10 &  bra, mm, todd, bonnie, ralph, casey, stacey, gordon, lou, dana
            
		\end{tabular}}
		
		\caption{Top-10 detected words from our method (NN) vs. AlignCos method (last row), for corpus split according to the gender of the tweet-author. Each word from our method is accompanied by its top-10 neighbors in each of the two corpora (Male vs. Female).} 
		\label{tab:gender_top10}
	\end{center}
	
\end{table*}

\begin{table*}
    
	\begin{center}
    	    \scalebox{0.75}{
    		\begin{tabular} {c||l}
    
            \multicolumn{2}{l}{Occupation (performer vs. sports)} \\\hline
            NN & neighbors in each corpus \\ \hline \hline
            \multirow{2}{*}{blues} &  funk, reggae, b.b., boneshakers, folk, bluegrass, grooves, rhythm, trippers, moody \\                 
            &  hawks, leafs, rangers, sabres, bruins, tahs, fulham, knights, yotes, maroons\\ \hline
            \multirow{2}{*}{cc} &  \includegraphics[scale=0.2]{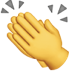}, \includegraphics[scale=0.2]{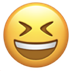},\includegraphics[scale=0.2]{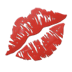},\includegraphics[scale=0.2]{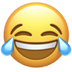} , \includegraphics[scale=0.2]{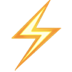}, \includegraphics[scale=0.2]{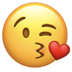}, \includegraphics[scale=0.1]{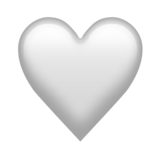}, \includegraphics[scale=0.2]{emojis/clap.png}, \includegraphics[scale=0.2]{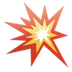}, lol \\                  
            &  sabathia, montclair, firestone, bethpage, isleworth, tourn, dorado, quail, riviera, westchester\\ \hline
            \multirow{2}{*}{dub} &  anime, subtitles, dubbing, dubbed, dlc, boxset, badman, rmx, miku, trax \\      
            &  lakeshow, crunk, yessir, w, yessirr, ayeeee, win, ayeeeee, yessirrr, yesir\\ \hline
            \multirow{2}{*}{bra} &  thong, panty, headband, panties, spanx, jeans, corset, uggs, tights, blouse \\
            &  bro, cuh, brodie, boi, dawg, brahh, breh, broo, cuzz, cuzo\\ \hline
            \multirow{2}{*}{track} &  rmx, tunes, album's, \includegraphics[scale=0.2]{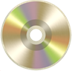}, trax, single, sampler, instrumental, unreleased, song's \\  &  racetrack, racin, field, slicks, velodrome, circuit, mtb, race, racing, sandown\\ \hline
            \multirow{2}{*}{wing} &  extremist, liberal, right-wing, fascists, leftist, conservative, propaganda, extremism, extremists, nationalists \\    &  flank, footed, rear, wingers, fullback, retake, netting, seat, midfield, fullbacks\\ \hline
            \multirow{2}{*}{par} &  ghar, nahin, dekhna, mujhe, rahe, kiya, apne, naam, aaj, theek \\                 &  pars, bogey, birdie, holes, putts, hole, putted, fairway, birdied, sawgrass\\ \hline
            \multirow{2}{*}{mo} &  starlite, reeds, knuckleheads, bossier, rocke, kcmo, stafford, granada, hutchinson, rosemont \\        
            &  bamba, tash, \includegraphics[scale=0.2]{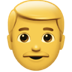}, wesley, kev, mane, yessssssssss, wes, yessssir, muzzy\\ \hline
            \multirow{2}{*}{ace} &  sweeeeet, fantastic, amazeballs, rad, amaaaazing, exceptional, sweeeet, jez, amazing-, hoot \\              
            &  sickkk, jb, robin, angel, stoner, ostrich, ayeeeee, milly, homey, hustler\\ \hline
            \multirow{2}{*}{duo} &  supergroup, violinist, troupe, cardenas, stylings, cellist, baritone, multi-talented, vocalist, bassist \\            
            &  tandem, northgate, dominant, keanu, hooker, wingers, rebounder, squads, superstar, jada\\ \hline \hline
            AlignCos Top-10 &  spencer, reed, dub, kurt, jerry, kirk, nova, watson, wa, curtis
            
		\end{tabular}}
		
		\caption{Top-10 detected words from our method (NN) vs. AlignCos method (last row), for corpus split according to the occupation of the tweet-author. Each word from our method is accompanied by its top-10 neighbors in each of the two corpora (performer vs. sports).} 
		\label{tab:occ_perf_sp_top10}
	\end{center}
	
\end{table*}

\begin{table*}
    
	\begin{center}
	    \scalebox{0.73}{
		\begin{tabular} {c||l}

            \multicolumn{2}{l}{Occupation (creator vs. sports)} \\\hline
            NN & neighbors in each corpus \\ \hline \hline
            \multirow{2}{*}{cc} &  \includegraphics[scale=0.2]{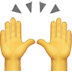}, \includegraphics[scale=0.2]{emojis/tearsofjoy.png},  \includegraphics[scale=0.2]{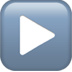},  \includegraphics[scale=0.2]{emojis/thinking.png},\includegraphics[scale=0.2]{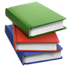}, xo-mk, rt, \includegraphics[scale=0.2]{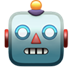}, \includegraphics[scale=0.2]{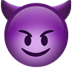}, \includegraphics[scale=0.2]{emojis/clap.png} \\       &  montclair, firestone, bethpage, isleworth, tourn, dorado, quail, riviera, westchester, vero \\ \hline
            \multirow{2}{*}{op} &  nel, reeva, roux, hoare, pathologist, shauna, baden-clay, ed, nedrow, barrister \\                   
            &  reconstruction, achilles, knee, ruptured, recovering, acl, surgeon, meniscus, tendon, injury\\ \hline
            \multirow{2}{*}{blues} &  reggae, bluegrass, fillmore, rhythm, rockers, ellington, grooves, techno, dnb, hob \\                   
            &  hawks, leafs, sabres, bruins, tahs, fulham, yotes, rovers, gunners, maroons\\ \hline
            \multirow{2}{*}{origin} &  ethnicity, ancestry, identity, significance, mythology, identification, protagonists, lineage, lore, retelling \\            
            &  nrl, afl, maroons, qld, footy, ashes, wallabies, a-league, premiership, roosters\\ \hline
            \multirow{2}{*}{wing} &  right-wing, far-left, faction, left-wing, zionist, reactionary, globalist, conservative, extremist, liberal \\                   &  flank, footed, fullback, retake, netting, seat, midfield, fullbacks, guard, mozzarella\\ \hline
            \multirow{2}{*}{weigh} &  meddle, defer, invest, bathe, reassure, implicated, experts, ponder, expel, summarize \\                   
            &  weigh-in, weigh-ins, ins, sparring, pre-fight, ufc, bellator, strikeforce, spar, ufcfightpass\\ \hline
            \multirow{2}{*}{worlds} &  universes, history's, colliding, realms, planets, universe, eras, modes, franchises, environments \\              
            &  europeans, olympics, worldcup, commonwealths, wc, commonwealth, championships, european, cwg, paralympics\\ \hline
            \multirow{2}{*}{sessions} &  comey, rosenstein, recusal, mcgahn, mccabe, recused, recuse, mueller, doj, dhs \\                  
            &  practices, sess, circuits, drills, weights, interval, camps, trainings, training, workout\\ \hline
            \multirow{2}{*}{track} &  rmx, compilation, reloaded, hexagon, soundcloud, ep, dnb, bandsintown, tunes, rework \\                   
            &  racetrack, racin, sx, field, slicks, velodrome, circuit, mtb, race, racing\\ \hline
            \multirow{2}{*}{presents} &  luts, voyager, housecall, ottaviani, uploaded, balearic, inharmony, derringer, machel, schulz \\                   
            &  pressies, pressie, advent, decorating, cupcakes, toys, x-mas, sweets, certificates, handmade\\ \hline \hline
            AlignCos Top-10 &  lawrence, marc, morris, op, diamond, carter, dash, cont, bee, norman

		\end{tabular}}
		
		\caption{Top-10 detected words from our method (NN) vs. AlignCos method (last row), for corpus split according to the occupation of the tweet-author. Each word from our method is accompanied by its top-10 neighbors in each of the two corpora (creator vs. sports).} 
		\label{tab:occ_cr_sp_top10}
	\end{center}
	
\end{table*}

\begin{table*}
    
	\begin{center}
	    \scalebox{0.75}{
		\begin{tabular} {c||l}

            \multicolumn{2}{l}{Occupation (creator vs. performer)} \\\hline
            NN & neighbors in each corpus \\ \hline \hline
            \multirow{2}{*}{echo} &  distortion, echoing, google's, lcd, ibooks, vibe, voice, songbook, audience, roku \\                   
            &  griffith, park, regents, acjokes, crest, roxy, paramount, trippers, folly, petco\\ \hline
            \multirow{2}{*}{inc} &  kopel's, acquires, takeover, selects, async, -short, sony, invests, blaqstarr, tata \\                  
            &  aimless, caa, phonte, psi, edu, morillo, fuentes, omega, intl, int'l\\ \hline
            \multirow{2}{*}{cont} &  rec, thru, mang, recs, mi, ul, sr, bsm, ing, tm \\ 
            &  thku, \includegraphics[scale=0.2]{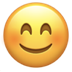}rt, oth, btw-, muah, 0)\includegraphics[scale=0.2]{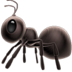}, vry, twd, \includegraphics[scale=0.2]{emojis/kiss.png}rt, wnt\\ \hline
            \multirow{2}{*}{presents} &  luts, voyager, housecall, ottaviani, uploaded, balearic, inharmony, derringer, machel, schulz \\                  
            &  morillo, erick, bash, pressies, whalum, pressie, winans, pawty, productions, torry\\ \hline
            \multirow{2}{*}{rebel} &  kurdish, libyan, jihadist, factions, sunni, jihadi, militant, hamas, daesh, isis \\                   
            &  ruler, rocker, geek, muse, whore, nerd, madonna's, daydream, gangster, hippie\\ \hline
            \multirow{2}{*}{buck} &  manziel, clayton, jerry, wiley, cowboys, romo, ambrose, flacco, kidd, mavs \\                   
            &  bucky, cocker, paperboy, rickie, hefner, mcdowell, roddy, cy, farmer, leadoff\\ \hline
            \multirow{2}{*}{thee} &  salute, paraphrase, bishop, esv, browning, faulkner, lia, medina, kaysha, atwood \\                  
            &  shalt, thyself, merciful, ephesians, hahahahahahah, thine, philippians, yesssssss, throne, humbly\\ \hline
            \multirow{2}{*}{chapter} &  prologue, prc, outlining, novella, pages, scene, heartstopper, cebu, tome, outline \\                   
            &  bl, tblst, sdmf, doom, grimmest, warhammer, quilt, draculas, dario, crusade\\ \hline
            \multirow{2}{*}{dash} &  jnr, peppermint, flashes, wop, keef, cappuccino, scotty, hummus, lily, disco \\                   
            &  skeetv, skee, radio, hbr, snip, twirl, \includegraphics[scale=0.2]{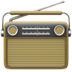}, blip, iheart, krispy\\ \hline
            \multirow{2}{*}{op} &  hoare, pathologist, shauna, baden-clay, nedrow, barrister, arguedas, protestor, bourque, arias \\                  
            &  urgentdogsofmiami's, doreenvirtue's, dermatologist, examination, surgeon, physio, intv, ons, nasal, doctor\\ \hline \hline
            AlignCos Top-10 & vince, todd, dana, watson, norman, marc, jerry, rs, mitch, brooks

		\end{tabular}}
		
		\caption{Top-10 detected words from our method (NN) vs. AlignCos method (last row), for corpus split according to the occupation of the tweet-author. Each word from our method is accompanied by its top-10 neighbors in each of the two corpora (creator vs. performer).} 
		\label{tab:occ_cr_pe_top10}
	\end{center}
	
\end{table*}

\begin{table*}
    
	\begin{center}
	    \scalebox{0.75}{
		\begin{tabular} {c||l}

            \multicolumn{2}{l}{Time of week (weekday vs. weekend)} \\\hline
            NN & neighbors in each corpus \\ \hline \hline
            \multirow{2}{*}{trick} &  sudoku, sneaky, summat, moonwalk, frighten, rubik's, clicker, smthng, stunt, foam \\                   
            &  treaters, treater, tricker, trick-or-treating, trick-or-treat, treats, or, neices, trick-or-treaters, kids\\ \hline
            \multirow{2}{*}{cc} &  citibank, debit, wachovia, credit, barter, visa, waived, payment, pkg, expedia \\                   
            &  snyder, ecu, rivera, mvc, yankees, clinches, natl, lin, ul, rk\\ \hline
            \multirow{2}{*}{ups} &  dhl, upping, situps, gowalla, shipment, shipments, fy, lunges, webos, sit-ups \\                   
            &  budgets, tractor, full-time, dri, radioshack, quik, distribution, fro, cheeseburgers, soulja\\ \hline
            \multirow{2}{*}{recall} &  recalling, maclaren, stork, cribs, strollers, defective, pedals, tundra, toyota, manufacturer \\                  
            &  fancy, specify, attribute, recommend, resist, adjust, vary, fwiw, grieve, refrain\\ \hline
            \multirow{2}{*}{rush} &  stampede, queues, detour, stretch, layover, standstill, congestion, levin, oncoming, braving \\                   
            &  refusal, jerry, pass, cbs, sellout, sideline, dover, interference, onside, tuscaloosa\\ \hline
            \multirow{2}{*}{bond} &  etfs, bernanke, insurer, sentencing, trustee, r.i., deficits, rba, hig, funds \\                   
            &  labor, humphrey, clarke, srk, titanic, fireman, colonel, fx, barney, jessie\\ \hline
            \multirow{2}{*}{pitch} &  bullpen, clinch, utley, win-win, lidge, interviewed, series, signage, stun, teleconference \\                  
            &  midfield, half-time, werth, tsn, offside, scoreless, roughing, punts, goal, rockies\\ \hline
            \multirow{2}{*}{lloyd} &  marv, asher, peter, andre, payton, phillip, bennett, o'connor, neal, wright \\                   
            &  llyod, jeward, mcelderry, lloyd's, ollie, stace, danyl's, jedwards, afro, olly's\\ \hline
            \multirow{2}{*}{zone} &  faction, wasteland, emp, vibin, i.e, l.a., constraints, realms, xtreme, jammin \\                   
            &  endzone, redzone, fumbled, fumbles, interceptions, touchdown, interference, bounds, interception, romo\\ \hline
            \multirow{2}{*}{ref} &  salary, overturn, statewide, applicants, amendments, position, ordinance, commissioning, nsw, anc \\                 
            &  offside, capello, burley, mangini, play-off, officiating, roughing, rooney, interference, fumbled\\ \hline \hline
            AlignCos Top-10 & maine, evan, griffin, terry, sp, aaron, ken, harris, todd, li
            
		\end{tabular}}
		
		\caption{Top-10 detected words from our method (NN) vs. AlignCos method (last row), for corpus split according to the time of week of the tweet. Each word from our method is accompanied by its top-10 neighbors in each of the two corpora (weekday vs. weekend).} 
		\label{tab:time_of_week_top10}
	\end{center}
	
\end{table*}

\begin{table*}
    
	\begin{center}
	    \scalebox{0.72}{
		\begin{tabular} {c||l}
		    \multicolumn{2}{l}{French (2014 vs. 2018)} \\\hline
            NN & neighbors in each corpus \\ \hline \hline
            \multirow{2}{*}{malcom} &  charmed, futurama, desperate, housewives, housewifes, simpson, hunter, ferb, smallville, scott \\                   &   dembele, coutinho, mariano, paulinho, rafinha, diakhaby, dembélé, dembelé, dembouz, rakitic\\ \hline
            \multirow{2}{*}{rn} &  en, eb, zn, en., enn, bored, bloquee, same, omfgg, stm \\                   &   fn, rn., dlf, fn., lfi, fhaine, lr, ex-fn, lrem, pcf\\ \hline
            \multirow{2}{*}{boe} &  bne, bnne, bonne, binne, bonnne, boonne, bone, bnn, bonnee, booonne \\                   &   peiffer, fourcade, svendsen, makarainen, schempp, desthieux, guigonnat, kuzmina, dahlmeier, tarjei\\ \hline
            \multirow{2}{*}{mina} &  kenza, ibtissem, bety, ghada, lina, laith, bzf, liya, ana, salom \\                   &  yerry, yerri, paulinho, gomes, mina., alcacer, rakitic, rafinha, dembele, coutinho\\ \hline
            \multirow{2}{*}{smet} &  smettre, smette, tmet, met, spose, senjaille, stappe, smettent, sdonne, samuse \\                   &   hallyday, laeticia, laura, læticia, vartan, halliday, hallyday., johnny, boudou, laetitia\\ \hline
            \multirow{2}{*}{lr} &  bdx, dk, poitiers, bx, rouen, caen, amiens, malo, perpi, aix \\                   &   lr., lrem, dlf, lfi, fn, ump, républicains, udi, vb, rn\\ \hline
            \multirow{2}{*}{divergent} &  tmr, tfios, thg, catching, hunger, mockingjay, fsog, insurgent, allegiant, tobias \\                   &   divergent., diverge, divergentes, diffèrent, convergent, diverger, diamétralement, concordent, opposées., divergences\\ \hline
            \multirow{2}{*}{ia} &  ya, y', yaura, quya, yavai, yaver, yora, yavait, yia, jconai \\                   &   artificielle, intelligenceartificielle, i.a, ia., intelligence, iot, i.a., artificielle., chatbots, automatisation\\ \hline
            \multirow{2}{*}{jdr} &  jdrr, hablais, duele, pfpfpfpfpf, eso, igual, nadie, déjame, pensar, pelis \\                   &   jdr., warhammer, shadowrun, roleplay, pathfinder, shmup, fangame, dungeon, rp, webcomic\\ \hline
            \multirow{2}{*}{cs} &  csst, ceest, enpls, wch, tst, cetei, wcch, c, ctei, cetai \\                   &   csgo, rl, pubg, fornite, fortnite, battlerite, faceit, ow, cod, dota\\ \hline
            AlignCos Top-10 & -l, malcom, maximilien, dna, lr, mina, boe, dias, sierra, giuseppe
		\end{tabular}}
		\caption{Top-10 detected words from our method (NN) vs. AlignCos method (last row), for corpus split according to the year of the text. Each word from our method is accompanied by its top-10 neighbors in each of the two corpora (2014 vs. 2018).}
		
		\label{tab:fr_top10}
	\end{center}
	
\end{table*}

\end{document}